\documentclass[10pt,twocolumn,letterpaper]{article}

\usepackage{cvpr}
\usepackage{times}
\usepackage{epsfig}
\usepackage{graphicx}
\usepackage{amsmath}
\usepackage{amssymb}

\usepackage{booktabs}       

\usepackage{multirow}

\usepackage[pagebackref=true,breaklinks=true,letterpaper=true,colorlinks,bookmarks=false]{hyperref}

 \cvprfinalcopy 

\def\cvprPaperID{6606} 

\ifcvprfinal\fi
\begin{document}

\title{SSRNet: Scalable 3D Surface Reconstruction Network}

\author{Zhenxing Mi \footnotemark[2] \and Yiming Luo \footnotemark[2] \and Wenbing Tao \footnotemark[1] \and 
	National Key Laboratory of Science and Technology on Multi-spectral Information Processing\\
	School of Artifical Intelligence and Automation, Huazhong University of Science and Technology, China \and
	{\tt\small \{m201772503, yiming\_luo, wenbingtao\}@hust.edu.cn} \\
}

\maketitle
\thispagestyle{empty}

\renewcommand{\thefootnote}{\fnsymbol{footnote}}

\footnotetext[2]{Equal contribution}
\footnotetext[1]{Corresponding author}

\renewcommand{\thefootnote}{\arabic{footnote}}

\begin{abstract}
Existing learning-based 
surface reconstruction
methods from point clouds  
are still facing challenges  
in terms of scalability and preservation 
of details on large-scale point clouds.
In this paper, we propose the SSRNet, 
a novel scalable learning-based method
for surface reconstruction. 
The proposed SSRNet constructs local geometry-aware  features for octree vertices and designs a scalable reconstruction pipeline, which not only greatly enhances the predication accuracy  of the relative position between the vertices and the implicit surface facilitating the surface reconstruction  quality, but also allows dividing the point cloud and octree vertices and processing different parts in parallel for superior scalability on large-scale point clouds with millions of points.   
Moreover, SSRNet demonstrates outstanding generalization capability and only needs several surface data for training, much less than other learning-based reconstruction methods, which can effectively avoid overfitting. The trained model of SSRNet on one dataset can be directly used on other datasets with superior performance.  
Finally, the time consumption with SSRNet on
a large-scale point cloud is acceptable and competitive.
To our knowledge, the proposed SSRNet is the first to really bring a convincing solution to the scalability issue of the learning-based surface reconstruction methods, and is an important step to make learning-based methods
competitive with respect to geometry processing  methods on real-world and challenging data.
Experiments show that our method achieves 
a breakthrough in scalability and quality compared with state-of-the-art learning-based 
methods. 
\end{abstract}

\section{Introduction}\label{sec:introduction}

\begin{figure}
	\begin{center}
		\includegraphics[width=\linewidth]{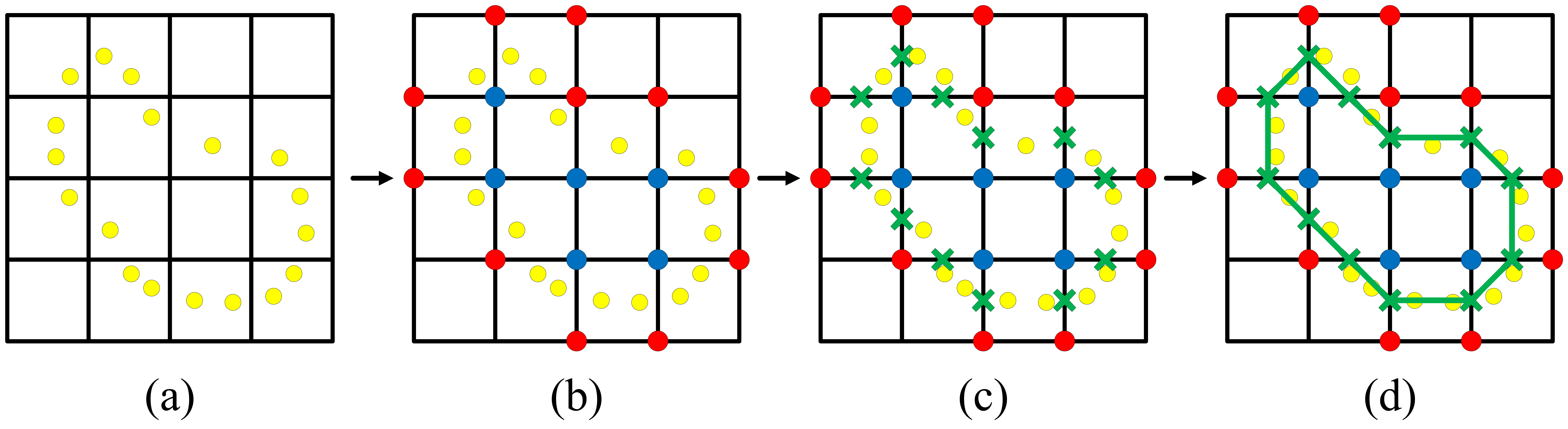}
	\end{center}
	\caption{2D example of surface reconstruction
		methods using implicit function.
		Yellow dots represent points in point cloud.
		Red dots represent vertices outside the implicit surface.
		Blue dots represent vertices inside the implicit surface.
		According to the implicit function value on vertices, 
		the MC method finds the intersections (green $ \bf{\times} $)
		of the grid lines and the implicit surface and then connects
		these intersections to extract a surface (green line).}
	\label{fig:marching_cubes}
\end{figure}

Point cloud is an important 
and widely used representation for 3D data. 
Surface reconstruction from point clouds (SRPC) 
has been well studied in computer graphics.
A lot of geometric reconstruction 
methods have been proposed
\cite{curless1996volumetric,carr2001reconstruction,turk2002modelling,levin2004mesh,kazhdan2006poisson,guennebaud2007algebraic,kazhdan2013screened,oltcheva2017surface}.
A commonly used pipeline 
in geometric reconstruction methods 
first computes an implicit function 
on a 3D grid
\cite{curless1996volumetric,kazhdan2013screened}.
Then the Marching Cubes (MC)  \cite{lorensen1987marching} 
is applied to extract an isosurface
from the 3D grid through implicit function values.
The grid is usually
an adaptive octree. The intersection of 
two grid lines in an octree is named as octree vertex.
Specifically, the implicit function can be an
indicator function that indicates 
whether the vertices are inside 
or outside the implicit surface. 
That is, a surface reconstruction 
problem can be seen as a binary 
classification problem for octree vertices
(See Figure \ref{fig:marching_cubes}).

In real-world reconstruction tasks, point clouds are
mainly acquired by scanners or Multi-View Stereo (MVS) methods
\cite{schonberger2016pixelwise,Xu_2019_CVPR}
 and are usually dense with more than millions of points and with complex topologies, 
which bring great challenges to surface reconstruction.
With the development of deep learning,  
a variety of learning-based 
surface reconstruction methods have been 
proposed
\cite{dai2017shape,riegler2017octnetfusion,
	liao2018deep,
	groueix2018papier,mescheder2019occupancy,
    park2019deepsdf}.
However, there are still problems for them to reconstruct real-world point clouds
for four reasons.

1) The network architectures
and output representations of these learning-based
methods usually force the learning-based methods to consider 
all the points at once.
Therefore, they can
not allow dividing input data 
and processing different parts separately. 
When point clouds scale to 
millions of points, devices may not be 
able to handle large amounts of memory usage,
which will make the model less scalable.
For example, the Deep Marching Cubes
\cite{liao2018deep} transforms the 
entire point cloud into voxel grids at once and directly outputs a 
irregular mesh, which is less scalable and only applicable to point clouds 
with thousands of points.

2) Although some methods can be applied to
large-scale point clouds, in reality they 
greatly downsample or aggregate the points
into a relative small scale.
For example, the Occupancy Networks (ONet)
\cite{mescheder2019occupancy}
and DeepSDF \cite{park2019deepsdf}
encode the entire points into a fixed-size
latent vector of 512 elements, which greatly limits
the representative power.
Consequently, they come at the expense
of surface quality reconstructed from large-scale 
point clouds.

\begin{figure}[t]
	\begin{center}
		\includegraphics[width=\linewidth]{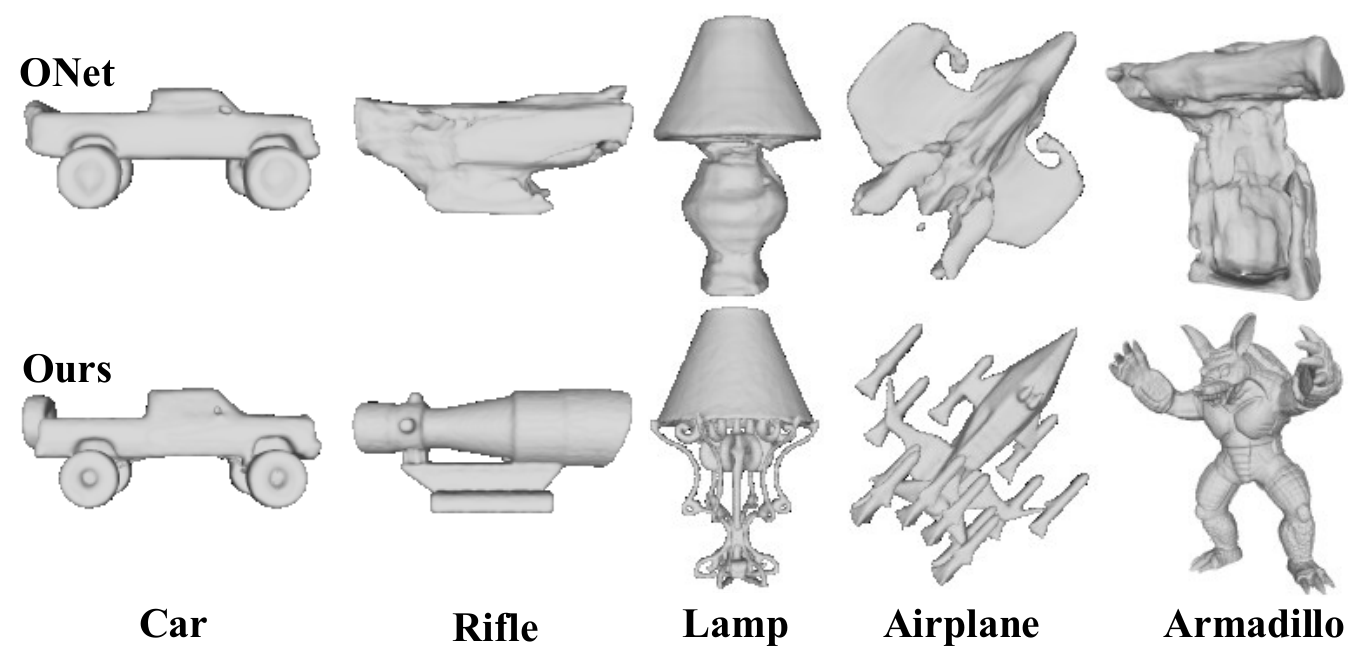}
	\end{center}
	\caption{Examples of reconstructed surfaces of
		ONet \cite{mescheder2019occupancy} and Ours on ShapeNet \cite{chang2015shapenet} and Stanford 3D  \cite{Stanford3D} (Armadillo).
	}
	\label{fig:intro_onet}
\end{figure}

3) Even on small-scale point clouds,
the existing learning-based methods are good at reconstructing 
shapes with quite simple topology rather than
those complex shapes.
In fact, the result surfaces of simple shapes 
are usually over smoothed.
The Figure \ref{fig:intro_onet} shows 
reconstruction examples
of a state-of-the-art 
learning-based method (ONet)
\cite{mescheder2019occupancy}
and ours.
For the simple shapes, ONet gets 
smooth surfaces. However, 
for the more complex topology shapes, 
it can not perfectly reconstruct the relevant details.

4) Some learning-based methods 
need a large portion of the dataset for training.
For example, ONet uses
4/5 of its dataset (about $ 40K $ 3D shapes) as training set.
Thus the network is inevitably more prone to over fitting
the dataset. What's more, it can not generalize well among different datasets, while SSRNet needs only several surface data for trainning and the trained model on one dataset can be directly used on the other datasets with superior performance.

%

In this paper, we take the advantages of 
the commonly used implicit function method
discussed above for surface reconstruction.
Specifically, the implicit function values
are 0-1 labels of octree vertices. 
They indicate whether the octree vertices are
in front or at back of the implicit surface.
The strategy simplifies
the surface reconstruction 
to binary classification 
problem, which is simpler and more 
effective. 
The ONet \cite{mescheder2019occupancy}
also trains its network in a binary
classification manner. 
However, this strategy does not 
guarantee high quality results.
There are three main limitations  in 
dealing with large-scale data that need to be broken.

1) \noindent{\textbf{The scalability of the network.}} 
A strong scalability of the classification network is important 
for tackling large-scale data. 
However, the commonly used PointNet \cite{qi2017pointnet} in reconstruction methods
considers entire points at once and 
is thus less scalable.
Our network does not 
need to input the entire data at once
 because the tangent convolution
\cite{tatarchenko2018tangent}
operations involved in our network
are all performed in a fixed-size local spatial region, 
and each $ 1 \times 1 $ convolution operation
is also independent. 
Moreover, these network operations 
are independent from the octree structures.
Therefore, our 
network allows dividing the points and 
octree vertices through bounding boxes and  
processing different parts separately
so that it wins strong scalability.

2) \noindent{\textbf{The ability to reconstruct geometry details.}} 
In order to achieve 
a high classification accuracy for octree vertices,
the ability to capture local geometry details 
is essential. 
Some methods like 
ONet\cite{mescheder2019occupancy} and DeepSDF \cite{park2019deepsdf} 
aggregate the features of the entire points into 
a fixed-size global latent vector (512 elements),
which lose a lot of geometry details.
This will inevitably lead to a 
decrease on reconstruction quality.
In our network, we focus on learning local
geometry information of the implicit surface.
Since the implicit surfaces are 
determined by octree vertex labels,
in order to make accurate classification 
for octree vertices, 
the most important task is to construct vertex features
properly.
The octree vertex features in our network 
are constructed from local neighbor points and 
take the advantage of local projection information, such as the signed projection distances
between octree vertices and 
their neighbor points, which are 
commonly used in geometric
methods such as the TSDF
\cite{curless1996volumetric}
method and MLS
methods \cite{levin2004mesh,guennebaud2007algebraic}.
They enable our method to 
capture local geometry details of 
large-scale point clouds and directly 
provide local geometry
information for 
accurate front-back classification.
Therefore, our network benefits 
from geometry-aware features 
to reconstruct accurate surfaces. 
Some other methods such as the 
ONet\cite{mescheder2019occupancy} 
and DeepSDF \cite{park2019deepsdf} 
construct the features of vertices 
by directly concatenating 3D 
coordinates of vertices with the
global latent vector 
of the point cloud.
This feature construction 
method cannot provide explicit 
information about relative  
position between vertices 
and the implicit surface.
Therefore, they cannot capture 
local geometry details well to 
reconstruct quality surfaces.


3) \noindent{\textbf{The generalization ability of the network.}}
Paying too much attention to global feature rather 
than just considering local information would limit 
the generalization performance among different types of shapes.
The octree vertex features in our network 
are local geometry-aware. 
They do not rely on global shape information excessively.
Therefore, it has good generalization capability
and does not need too much training data, 
which avoids overfitting on dataset.

Overall, our contributions can be summarized as follows.

\begin{itemize}
	\item We design a scalable pipeline
	for reconstructing surface from 
	real-world point clouds.
	\item We 
	construct local geometry-aware 
	octree vertex feature, which leads
	to accurate classification for octree
	vertices and good generalization capability 
	among different datasets.
\end{itemize}

Experiments have shown that our method achieves 
a significant improvement
in scalability and quality compared
with state-of-the-art learning-based 
methods, and is competitive with respect to  state-of-the-art 
geometric processing methods in terms of reconstruction quality and efficiency
\footnote{Our code will be available in Github later.}.

\section{Related Work}\label{sec:relatedwork}

In this section, we first review some 
important geometric reconstruction methods
to introduce basic concepts.
Then we focus on existing learning-based surface reconstruction methods. We mainly analyze whether 
they are able to
scale to large-scale datasets and to 
capture geometry details in terms of network architectures and 
output representations.


\subsection{Geometric reconstruction methods}

Geometric reconstruction methods 
can be broadly categorized into 
global methods and local methods.
The global methods consider all the data at once, 
such as the radial basis functions (RBFs) methods
\cite{carr2001reconstruction,turk2002modelling},
and the (Screened) Poisson Surface Reconstruction method (PSR)
\cite{kazhdan2006poisson,kazhdan2013screened}.

The local fitting methods usually define 
a truncated signed distance function 
(TSDF) \cite{curless1996volumetric} 
on a volumetric grid. 
The various moving least squares methods (MLS) 
fit a local distance field or
vector fields by spatially varying low-degree polynomials, 
and blend several 
nearby points together 
\cite{levin2004mesh,guennebaud2007algebraic}.
The local projection 
and local least squares fitting used by MLS
are similar to the tangent convolution 
used in our network.
Local methods can be well scaled. 

There are also other geometric methods with 
different strategies, such as \cite{oltcheva2017surface}
computing restricted Voronoi
cells. More detailed reviews and evaluations
of geometric methods can be found 
in \cite{berger2014state} and \cite{zhu2019multisource}.


\subsection{Learning-based reconstruction methods}

\noindent{\textbf{Network architecture.}} 
One straightforward network 
architecture for point clouds 
is to convert the point clouds 
into regular 3D voxel grids or adaptive 
octree grids and apply 3D convolutions
\cite{maturana2015voxnet,riegler2017octnet}.
The voxel-based network used in 3D-EPN  \cite{dai2017shape} 
and the Deep Marching Cubes (DMC)
 \cite{liao2018deep}  faces cubic growth
of computation and memory requirements for large-scale datasets.
The OctNetFusion 
\cite{riegler2017octnetfusion} and 
3D-CFCN \cite{cao2018learning}
reconstruct implicit surface 
from multiple depth images based on OctNet
\cite{riegler2017octnet}, a octree-based network, whose operations  are complicated and 
are highly related to the octree structure. Therefore,
they also face computational issues  in reconstructing
 large-scale datasets.

Another type of network architecture 
learns point features directly.
The commonly used point cloud network 
is PointNet \cite{qi2017pointnet}. 
It encodes global shape features
into a latent vector of fixed size.
Some reconstruction
methods also extract a latent vector from 
the point cloud, such as
ONet
\cite{mescheder2019occupancy} 
and DeepSDF \cite{park2019deepsdf}.
These networks 
are able to encode
features of a large-scale 
point cloud into a latent vector.
However, the latent vector 
of a small size limits
its representative power for
complex point clouds.

There are also network architectures learning
local features of point clouds. 
PointNet++ \cite{qi2017pointnet++}
groups points into overlapping parts, 
then PointNet is chosen as the local 
feature learner for each part.
The centroids of each parts are selected 
using iterative farthest point sampling (FPS). 
The points are grouped around the centroids by ball query 
that finds all points 
within a radius to the query point.
Using FPS with grouping operations cannot be  
guaranteed to be performed
 in a 
fixed-size spatial region.
Therefore, the PointNet++ does not allow
dividing input points 
using bounding boxes and processing each part 
independently.
The tangent convolution network
\cite{tatarchenko2018tangent}
learns local features from neighbor 
points
for semantic segmentation 
of 3D point clouds.
It defines
three new network operations:
tangent convolution, pooling and unpooling.
The neighbor points 
are collected by ball query.
The pooling and unpooling are
implemented via hashing onto a regular 3D grid.
These operations are performed in a fixed-size spatial region.
Therefore, the tangent convolution network
can divide points using bounding boxes and process
each part independently.

\noindent{\textbf{Output representation.}}
The scalability of learning-based 
reconstruction methods
is also greatly influenced by the output 
representation.
The reconstruction methods based on voxel or octree 
grids usually use the occupancy or TSDF on grids
as the output representation. 
This output representation, together 
with their network operations, is highly 
related to the grid structure.
Therefore, their scalability is limited.
The Deep Marching Cubes
 \cite{liao2018deep} and 
 AtlasNet 
\cite{groueix2018papier} directly 
produce a triangular surface.
The predictions of different parts of 
the surface are interdependent
so they have to consider the
whole input point cloud at once.

ONet
\cite{mescheder2019occupancy} and
 DeepSDF \cite{park2019deepsdf}  
learn shape-conditioned classifiers
whose decision boundary is the surface.
Their representations are similar to the front-back 
representation of our SSRNet.
Although ONet and DeepSDF 
support to make predictions over 3D locations parallelly,
they need to get the global 
latent vector of the whole point cloud 
so they do not allow 
dividing input points. 




\begin{figure}
	\begin{center}
		\includegraphics[width=\linewidth]{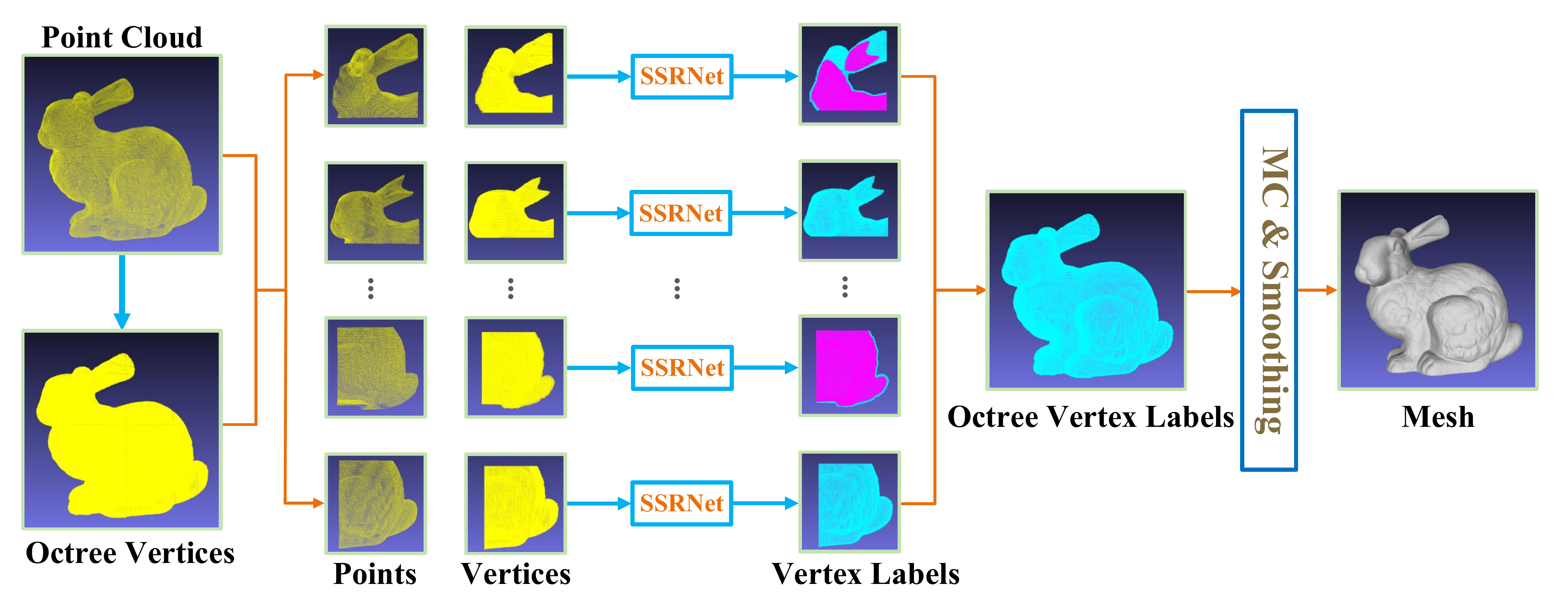}
	\end{center}
	\caption{The pipeline of our method. 
		Different parts are processed by our SSRNet parallelly.
	}
	\label{fig:pipeline}
\end{figure}

\section{Method}

\subsection{Pipeline}\label{sec:overviewoftsr}

We design a scalable learning-based 
surface reconstruction method, named SSRNet,
which allows dividing the input point clouds 
and octree vertices.
Figure \ref{fig:pipeline} shows the pipeline of our method.
Let $ \mathcal{P}=\{\textbf{p}_{i}|i=1, ..., N\} $ be the 
input point cloud of $ N $ points,
where $ \textbf{p}_{i} $ is point coordinate 
with normal $ \textbf{n}_{i} $.
We construct an octree from 
this point cloud and extract 
the vertices $ \mathcal{V} = \{\textbf{v}_{i}|i=1, ..., M\} $
of the finest level of this octree,
where each vertex $ \textbf{v}_{i} $ is 
the coordinate. 
To prevent the whole input data exceeding the GPU memory,
we use bounding boxes to divide
the point cloud and the vertices 
into $ \{\mathcal{P}_{j}|j=1, ..., K\} $ and
$ \{\mathcal{V}_{j}|j=1, ..., K\} $, where
$ K $ is the number of boxes.
The bounding boxes of vertices are not 
overlapping. The corresponding bounding boxes
of points are larger than those of octree vertices,
which ensures that all the neighbor points 
needed by vertices are included in the input. 
The border effects are thus largely suppressed. 
This enables our network to make accurate predictions
for vertices in the boundary of the bounding box.
We do not see any decay of accuracy on the borders of vertex boxes.
For the part $ j $, the vertices $ \mathcal{V}_{j} $
and the corresponding points $ \mathcal{P}_{j} $ 
are fed into SSRNet.
The SSRNet classifies each vertex $ \textbf{v}_{ji} $ 
in $ \mathcal{V}_{j} $ 
as in front or at back of the
implicit surface represented by
$ \mathcal{P}_{j} $. 
Let the function represented by the network
be $ f_\theta $, where 
$ \theta $ represents the parameters of the network.
Then the key  of our 
method is to classify $ \textbf{v}_{ji} $ in $ \mathcal{V}_{j} $.
It can be formulated as:
\begin{equation}
	f_{\theta}(\mathcal{P}_{j}, \mathcal{V}_{j}, \textbf{v}_{ji}) \rightarrow \{0, 1\}
\end{equation}

The front or back of an vertex is
defined by the normal direction
of its nearest surface.
The vertices are all from the 
finest-level voxels of the octree 
in order to reconstruct more details.
They do not contain vertices far from 
the actual surface, which brings great challenges
in classifying the vertices.
%
It is worth noting that the network operations
in SSRNet are not related to the structure of the octree.
After all the vertices are labeled, 
we extract surface using Marching Cubes (MC)
and post-process the surface with a simple 
Laplacian-based mesh smoothing method.


\subsection{Geometry-aware Vertex Feature}\label{sec:geometryawarevertexfeature}

As discussed in the Introduction,
the most important features
for octree vertex classification
are signed projection distances 
among octree vertices and 
their neighbor points.
In order to get accurate classification
for octree vertices and good generalization
capability, 
we design geometry-aware 
features for octree vertices directly encoding local 
projection information.

The tangent convolution 
\cite{tatarchenko2018tangent} is defined as 
the 2D convolution on tangent images which are constructed by local projections of
neighbor points.
The indices of the projected points 
used for tangent image can be precomputed,
which makes tangent convolution 
much more efficient.
The signals used in tangent images 
represent local 
surface geometry, including signed projection distances,
normals, etc. 
Therefore, it
has the potential to encode local geometry
features for octree vertices.

However, there are problems in
applying tangent convolution 
to octree vertices directly.
For a 3D location $ \mathbf{p} $,
the normal 
$\mathbf{n}_{\mathbf{p}}$
of its tangent image
is estimated through local covariance
analysis \cite{salti2014shot}. 
Let $\mathbf{k}$ be the eigenvector 
related to the smallest eigenvalue 
of the covariance matrix of $ \mathbf{p} $. The
normal of tangent image is defined as $\mathbf{k}$.
Due to the direction of eigenvector $\mathbf{k}$
is ambiguous,
it may not
be consistent with the real
orientations of local 
implicit surfaces. 
The signs of projection distances 
hence do not represent 
front-back information
accurately.

In order to solve this problem, 
we modify the definition
of $\mathbf{n}_{\mathbf{p}}$.
Since the front-back classification is 
related to neighbor points,
we use the input normals of neighbor 
points as additional 
constraints to define $\mathbf{n}_{\mathbf{p}}$.
Let $\mathbf{n}_{a}$ be the 
average input normal of the neighbor 
points of $\mathbf{p}$.
In our definition, if the angle 
between eigenvector $\mathbf{k}$ and $\mathbf{n}_{a}$ 
is more than $90^{\circ}$, we invert 
the direction of $\mathbf{k}$. 
Then we use it as the normal $\mathbf{n}_{\mathbf{p}}$ 
of the tangent 
image. 
That is, our definition of the tangent image ensures that
$\mathbf{n}_{\mathbf{p}}^{\top} \mathbf{n}_{a}>0$.

The features of octree vertices constructed
by modified tangent convolution directly
encode front-back information.
They are local geometric features and not related to
global information of shapes, so our network
is scalable and can generalize well among
different datasets. 
It's worth noting that we use 
neighbor points in the point cloud 
rather than neighbor vertices to compute the 
tangent images. It is because our network classifies
octree vertices with respect to the surface represented
by points from $\mathcal{P}$, rather than
being represented by neighbor vertices.

\subsection{Network}\label{sec:networkintroduction}

\begin{figure}
	\begin{center}
		\includegraphics[width=\linewidth]{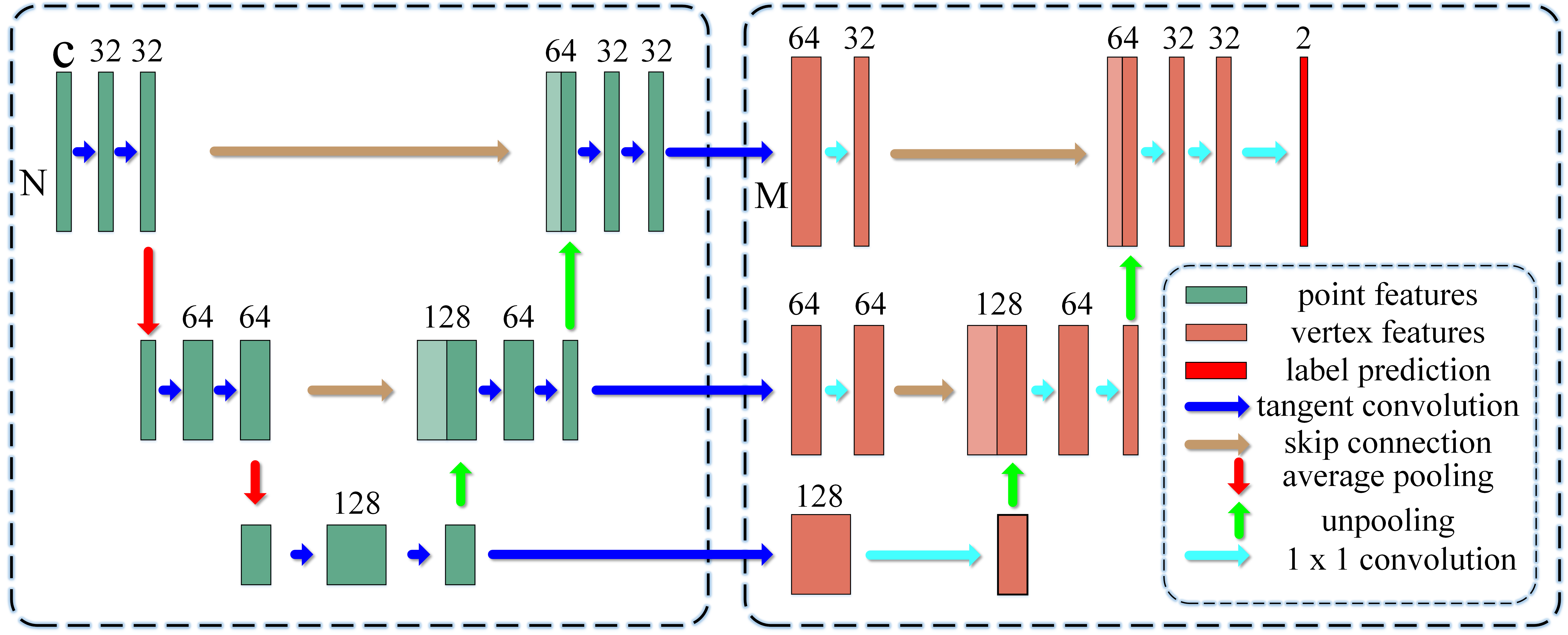}
	\end{center}
	\caption{Our network architecture. 
		Arrows of different colors represent different network operations.
		The inputs of this network include $ N $ points and $ M $ octree vertices.
		The initial features of the points have $ C $ channels.}
	\label{fig:networkarch}
\end{figure}

\noindent{\textbf{Network architecture.}}
The network architecture of our SSRNet
is illustrated in Figure \ref{fig:networkarch}.
It contains two parts. The left part is for
point feature extraction. The right 
is for vertex feature construction and label prediction.
They are connected by tangent convolution layers 
for vertices.

The left part of our network 
encodes the features of 
the input  $ N $ points
from a 
point cloud. 
It is a fully-convolutional 
U-shaped network with skip connections.
It is built by our 
modified tangent convolution layers.
Two pooling operations are used to increase the size of
the receptive field.
The indices of the projected points 
used for tangent images and the index 
matrices used by pooling and unpooling
layers are precomputed before the data are fed 
into the networks.

The right part is the core of our method. 
The geometry-aware
features of octree vertices 
are firstly
constructed by
modified tangent convolution, from point features 
in corresponding 
scale levels.
The indices of neighbors points from 
point cloud for tangent images of vertices
are also precomputed before network operations.
Then $1 \times 1$ 
convolutions and unpooling operations are applied to
encode more abstract features and to
predict labels for each vertex.
The index matrices used by unpooling
layers in this part are precomputed
when downsampling vertices via grid hashing. 


The input signals used by our method are all local signals.
We use signed distances from 
neighbor points to the tangent plane (D), 
and normals relative to the normal of the tangent
plane (N) as input. 
Since the network operations in SSRNet are all performed in a fixed-size local region,
i.e. the modified tangent convolution, 
the $1 \times 1$ 
convolution, the pooling and unpooling
of points and vertices,
our network allows dividing the points and vertices
with bounding-boxes and processing each part separately.
It is worth noting that bounding boxes used for points
are larger than those of octree vertices in order to make accurate
predictions for vertices near the boundary.

\noindent{\textbf{Implementation details.}}
In our implementation, 
we downsample the input points and 
vertices using grid hashing for pooling 
and unpooling.
Given an octree of depth $ D $, 
The length of the finest-level edges
is $ 1/2^{D} $.
We set the initial receptive field 
size in our network as $ 4/2^{D} $.
The grid hashing size of input 
points at three scales are set as 
$ 1/2^{(D+2)} $, 
$ 1/2^{(D+1)} $,
$ 1/2^{D} $.
We do not apply gird hashing for vertices 
in the first scale because the vertices 
have an initial grid size of $ 1/2^{D} $ originally.
The last two grid hashing sizes of vertices are set as
$ 1/2^{(D-1)} $
and $ 1/2^{(D-2)} $.
In order to retain more surface details
for the feature construction of vertices, 
we set smaller grid 
hashing sizes for points.

\subsection{Surface Extraction and Smoothing}\label{sec:surfaceextraction}

We use MC
 \cite{lorensen1987marching} 
to extract surface from the labeled octree vertices.
MC finds the intersections 
between octree edges and
the implicit surface using the labels of vertices.
Since we directly use the mid-points
of edges as the intersections, the resulting 
mesh has discretization artifacts inevitably.
We use a simple Laplacian-based smoothing method 
\cite{taubin1995signal}
as a post-processing step to refine the mesh.

\subsection{Data Preparation}\label{sec:datapreparation}

To prepare training data, we first normalize the coordinates of 
the input point cloud and surfaces. 
Then an octree is built from the point cloud. 
The octree building method is adapted from
the open-source code 
of PSR
\cite{kazhdan2013screened}. 
It ensures that the finest level
of the result octree contains
not only cubes with points in them,
but also their $3 \times 3 \times 3$ neighbor cubes.
Therefore, the octree is dense enough to completely reconstruct
the surface. 
We then use the ground truth surface with normals to 
label these vertices.
For datasets without ground truth surfaces, we 
generate ground truth surfaces through 
PSR.
More detailed introduction of our 
data preparing method can be found in our
supplementary file.

\begin{figure*}[t]
	\begin{center}
		\includegraphics[width=\linewidth]{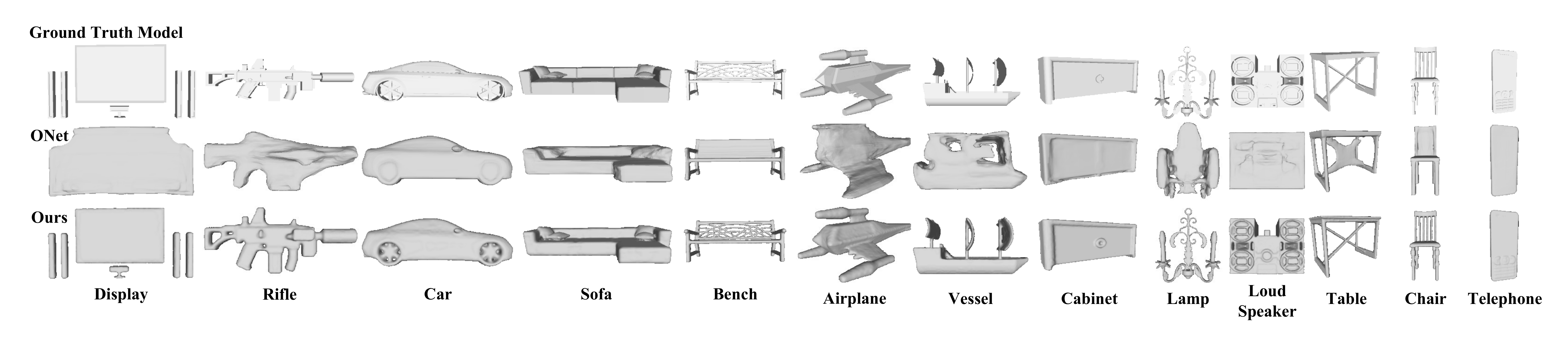}
	\end{center}
	\caption{Examples of reconstructed surfaces on testing set of 
		ShapeNet \cite{chang2015shapenet} by ONet \cite{mescheder2019occupancy} and our method.}
	\label{fig:shapenet_quality}
\end{figure*}

\section{Experiments}

In this section we perform
a series of experiments
on datasets of different scales to evaluate our method.
The datasets are all widely used in 3D-related tasks. 
We mainly examine the ability to capture 
geometry details, the scalability
and generalization capability of
our method on both small-scale and large-scale
datasets.
Due to the lack of triangular meshes,
the ground-truth meshes
used for training
are all produced by 
PSR.
It is worth noting that we just get ground-truth labels from the
triangular surfaces produced by PSR in the experiments due
to the lack of public available large-scale surface datasets.
However, our labels are not only available from PSR.

\subsection{Datasets and Evaluation Metrics}\label{sec:datasetsandevaluationmetrics}
\noindent{\textbf{Datasets.}} 
In order to compare with several 
state-of-the-art learning-based methods,
%
we do experiments on the same subset of ShapeNet 
\cite{chang2015shapenet} as 
ONet \cite{mescheder2019occupancy}, 
in which the point clouds are relatively small with 
tens of thousands of points in each.
Furthermore, 
in order to better evaluate the scalability 
and generalization capability to 
handle large-scale point clouds,
we choose DTU 
dataset \cite{jensen2014large}
and Stanford 3D Scanning Repository \cite{Stanford3D}.
Most point clouds in the two datasets contain millions of points. 

\noindent{\textbf{Evaluation Metrics.}} 
The direct metric of our network 
is the classification accuracy of octree vertices. 
It is worth mentioning that the vertices in our experiments are all from the finest-level voxels of the octree.
With
octree of relatively high resolution, these voxels do not contain vertices very far from the actual surface.  
Therefore, a high classification accuracy 
matters a lot to the quality of the final mesh.
For mesh comparison on ShapeNet,
we use the same evaluation metrics with 
ONet,
including the volumetric IoU (higher is better), the
Chamfer-$ L_1 $ distance (lower is better) and the normal consistency score (higher is better).
On the DTU dataset, we use DTU evaluation 
method \cite{jensen2014large}, 
which mainly evaluates DTU Accuracy 
and DTU Completeness (both lower is better).
And for Stanford datasets, Chamfer distance (CD) (lower is better) 
is adopted in order to evaluate the points of the output meshes, 
and the CD is also taken into consideration in the evaluation on DTU 
dataset. For each point in a cloud, CD finds the nearest
point in the other point cloud, and averages the square of
distances up.


\begin{table}[t]
	\centering
	\caption{Classification Accuracy of SSRNet on Datasets of different scales. ShapeNet-Model and DTU-Model represent the models we trained on ShapeNet \cite{chang2015shapenet} and DTU \cite{jensen2014large} respectively.}
	\resizebox{\columnwidth}{!}{
		\begin{tabular}{ccccc}
			\toprule
			\multirow{2}{*}{Dataset}&  \multirow{2}{*}{ShapeNet \cite{chang2015shapenet}} & \multirow{2}{*}{DTU \cite{jensen2014large}} & \multicolumn{2}{c}{Stanford \cite{Stanford3D}}                 \\
			\cmidrule(r){4-5}
			& & &  ShapeNet-Model &  DTU-Model   \\
			\midrule
			Accuracy (\%) & 97.6 & 95.7 & 98.1 & 98.2  \\
			\bottomrule
	\end{tabular}}
	\label{tab:classification_accuracy_of_SSRNet_on_different_scale_datasets}
	
\end{table}

\subsection{Results on Shapenet}\label{sec:resultsonshapenet}

In our first experiment, we evaluate the capacity of 
our network to capture shape details on ShapeNet.
We use the same test split as 
ONet
for fair comparisons.
The testing set contains
about $ 10K $ shapes.
We only randomly select a 
subset from the training set of ONet
for training.
The training set of our 
method ($ 4K $ shapes) is 1/10 
of the training set of ONet ($ 40K $ shapes).
We do not use a large amount of training data 
as ONet did mainly for two reasons. 
On the one hand, it is necessary to eliminate the 
possibility of overfitting a dataset due to too much training data.
On the other 
hand, our network learns local geometry-aware 
features so that it can capture
more geometry details with less training data.
Therefore, it is unnecessary for us 
to use too many training samples.

In our experiments, 
each point cloud in training and 
testing set contains $ 100K $ 
points. 
We add Gaussian noise with 
zero mean and standard 
deviation 0.05 to the
point clouds as ONet did. 
Our network is efficient at processing point clouds with a large number of points, so we do not downsample the point clouds.
We think it is important for Surface Reconstruction from Point Clouds (SRPC)
methods to be adaptive to the original data, rather than reducing the amount of input data and the resolution to fit their
networks. After all, the representative power of 300 points
used by ONet
is really limited. 
The original synthetic models of ShapeNet
do not have consistent normals
so that we reconstruct surfaces 
from the ground truth point clouds 
using PSR on octrees of 
depth 9 to generate the training data.
We use octrees of depth 
8 in SSRNet for 
training and testing.

For mesh comparison,
we use the evaluation metrics 
on ShapeNet mentioned in Section 
\ref{sec:datasetsandevaluationmetrics},
including the volumetric IoU, the
Chamfer-$ L_1 $ distance 
and the normal consistency score.
We evaluate all shapes in the testing set on these 
metrics.
The results of existing 
learning based 3D reconstruction approaches, 
i.e. the 3D-R2N2 \cite{choy20163d}, 
PSGN \cite{fan2017point}, 
Deep Marching Cubes (DMC) 
\cite{liao2018deep}
and ONet \cite{mescheder2019occupancy}, 
are obtained from the paper of ONet. 
As mentioned in ONet,  
it is not possible to evaluate 
the IoU for PSGN for the reason 
that it does not yield watertight meshes. 
ONet adapted 3D-R2N2 and
PSGN for point cloud
input by changing the encoders. Although methods in Table
\ref{tab:results_on_ShapeNet} use different strategies, they actually solve the same task
of SRPC. 
Quantitative results are shown
in 
Table \ref{tab:results_on_ShapeNet}.

\begin{table}[t]
	\centering
	\caption{Results of learning-based methods on ShapeNet \cite{chang2015shapenet}. 
		NC = Normal Consistency. The volumetric IoU (higher is better), the
		Chamfer-$ L_1 $ distance (lower is better) and NC (higher is better) are reported.}
	\resizebox{0.75\columnwidth}{!}{
		\begin{tabular}{cccccc}
			\toprule
			& IoU & Chamfer-$ L_1 $ & NC \\
			\midrule
			3D-R2N2 \cite{choy20163d} & 0.565 & 0.169 & 0.719  \\
			PSGN \cite{fan2017point}    & --   & 0.202 & --    \\
			DMC \cite{liao2018deep}     &  0.647 & 0.117 & 0.848    \\
			ONet \cite{mescheder2019occupancy}    & 0.778 & 0.079 & 0.895  \\
			Ours    & \textbf{0.957} & \textbf{0.024} & \textbf{0.967}  \\
			\bottomrule
	\end{tabular}}
	\label{tab:results_on_ShapeNet}
	
\end{table}

The classification accuracy of octree
vertices on ShapeNet dataset shows strong robustness of our network. 
It achieves high classification accuracy of $ 97.6\% $ 
on noisy point clouds (See Table \ref{tab:classification_accuracy_of_SSRNet_on_different_scale_datasets}). 
The Table \ref{tab:results_on_ShapeNet} 
shows that our method gets great
improvements on these metrics.
Compared with the best performance of other methods,
we achieve the highest IoU, 
lower Chamfer-$ L_1 $ distance 
and higher normal consistency.
More specifically, the IoU of our results is
about 18\% higher than that of ONet.
It is worth noting that our network has only 
0.49 million parameters, while ONet 
has about 13.4 million parameters.

As shown in Figure \ref{fig:shapenet_quality},
the surface quality of ours is generally 
better than those of ONet.   
Our network 
is good at capturing local geometry details
of shapes of different classes, 
even with quite different 
and complex topology.
ONet can reconstruct smooth 
surfaces for shapes with simple 
topology. However, since the global 
latent vector encoder in ONet loses 
shape details, it tends to generate 
an over-smoothed shape for a complex 
topology. 
For more visual results, please see our supplementary materials.

%
%
%
%
\begin{table}[t]
	
	\centering
	\caption{Surface quality on DTU \cite{jensen2014large} testing
		scenes. Surfaces used for evaluation
		criterias are all reconstructed at octree depth 9. DA=DTU
		Accuracy, DC=DTU Completeness, CD=Chamfer distance (all lower is better).}
	\resizebox{\columnwidth}{!}{
		\begin{tabular}{cccccccccc}
			\toprule
			\multirow{2}{*}{Method}  &   \multicolumn{2}{c}{DA} & \multicolumn{2}{c}{DC} & \multicolumn{2}{c}{CD}                 \\
			\cmidrule(r){2-3} \cmidrule(r){4-5} \cmidrule(r){6-7}
			& Mean & Var. & Mean & Var. & Mean & RMS   \\
			\midrule
			PSR(trim 8) \cite{kazhdan2013screened}    & 0.473    & 1.33    & 0.327    & 0.220          & 3.16 & 12.5  \\
			PSR(trim 9.5) \cite{kazhdan2013screened}  & 0.330    & 0.441    & 0.345    & 0.438         & \textbf{1.17} & 4.49  \\
			Ours & \textbf{0.321} & \textbf{0.285} & \textbf{0.304} & \textbf{0.0888} & 1.46 & \textbf{4.42} \\
			\bottomrule
	\end{tabular}}
	\label{tab:dtu_accu}
	
\end{table}

\begin{table}[t]
	\centering
	\caption{Generalization on Stanford 3D \cite{Stanford3D}.The Chamfer Distances 
		are in units of $ 10^{-6} $. Surfaces used for distance criterias are all reconstructed at octree depth 9. Ours-S and Ours-D represent the models we trained on ShapeNet \cite{chang2015shapenet} and DTU \cite{jensen2014large} respectively.}
	\resizebox{\columnwidth}{!}{
		\begin{tabular}{ccccccccc}
			\toprule
			\multirow{3}{*}{Data}  & \multicolumn{2}{c}{Accuracy(\%)} & \multicolumn{3}{c}{CD Mean} & \multicolumn{3}{c}{CD RMS}                  \\
			\cmidrule(r){2-3} \cmidrule(r){4-6} \cmidrule(r){7-9} 
			& Ours & Ours & \multirow{2}{*}{ONet \cite{mescheder2019occupancy}} & Ours & Ours & \multirow{2}{*}{ONet \cite{mescheder2019occupancy}} & Ours & Ours \\
			& S & D &  & S & D &  & S & D \\
			\midrule
			Armadillo  & 98.2 & 97.8 & 93.46 & 0.028 & \textbf{0.023} & 168.59 & 0.131 & \textbf{0.056} \\
			Bunny      & 98.2 & 98.7 & 94.88 & 0.064 & \textbf{0.057} & 165.44 & 0.134 & \textbf{0.087} \\
			Dragon     & 98.0 & 98.0 & 40.69 & 0.053 & \textbf{0.047} & 74.99 & 0.208 & \textbf{0.150} \\
			\bottomrule
	\end{tabular}}
	\label{tab:generalization_on_stanford}
	
\end{table}

\begin{table}[t]
	\centering
	\caption{The time performance of our method. We report 
		the first scale point number, first scale octree 
		vertex number and triangle number in our results
		using million (M) as unit.
		The preprocessing time (Prep. time) includes octree construction time, 
		downsampling time, tangent images precomputing time and batches computing time.
		The prediction time (Pred. time) is the time for loading partitioned points 
		and vertices and predicting labels. }
	\resizebox{\columnwidth}{!}{
		\begin{tabular}{cccccccccccc}
			\toprule
			Number & Armadillo & Bunny & Dragon & stl\_030 & stl\_034 & stl\_062 \\ 
			\midrule
			Point /M                  & 2.16  & 0.361  & 1.71  & 2.43  & 2.01   & 2.19   \\ 
			Vertex /M               & 3.29  & 3.62   & 3.07  & 1.07  & 0.766  & 0.922  \\ 
			Triangle  /M                   & 1.18  & 1.52   & 1.16  & 0.42  & 0.31   & 0.36   \\ 
			Batch                     & 109   & 73     & 77    & 59    & 88     & 86     \\ 
			
			\bottomrule
			\toprule
			
			Time /s & Armadillo & Bunny & Dragon & stl\_030 & stl\_034 & stl\_062 \\ 
			\midrule
			Prep               & 19.4  & 8.86   & 16.6  & 10.9  & 9.34   & 9.99   \\ 
			Pred (1 GPU)   & 133   & 82.2   & 110.5 & 63.5  & 84.5   & 87.9    \\ 
			Pred (4 GPUs)  & \textbf{50.5}  & \textbf{27.3}   & \textbf{38.2}  & \textbf{30.4}  & \textbf{31.1}   & \textbf{34.3}    \\ 
			\midrule
			Total (1 GPU)        & 153   & 92.0   & 128   & 75.4  & 94.5   & 98.6    \\ 
			Total (4 GPUs)        & \textbf{70.6}  & \textbf{37.1 }  & \textbf{55.7}  & \textbf{42.3}  & \textbf{41.1}   & \textbf{45.0}    \\ 
			
			\bottomrule
	\end{tabular}}\label{tab:total_time}
	
\end{table}

\begin{figure*}[t]
	\begin{center}
		\includegraphics[width=\linewidth]{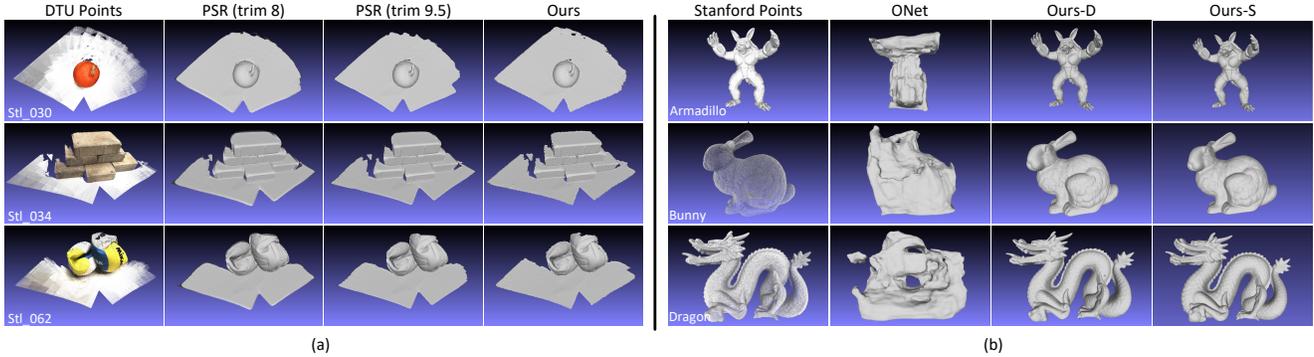}
	\end{center}
	\caption{Examples of surfaces of our method and PSR \cite{kazhdan2013screened} on DTU \cite{jensen2014large} (a).
		Generalization testing results of our method and ONet \cite{mescheder2019occupancy} 
		on Stanford 3D \cite{Stanford3D} (b).
Ours-S and Ours-D represent the our models trained on ShapeNet \cite{chang2015shapenet} and DTU \cite{jensen2014large} respectively.
		Background colors are 
		for better visualization for point clouds.
	}
	\label{fig:dtu_quality}
\end{figure*}

%
%

\subsection{Results on 3D Scans of Larger Scales}\label{sec:comparepsr}


\noindent{\textbf{Evaluation on DTU dataset.}
We train and test our network on DTU at octree depth 9.
Since the ground truth surfaces of DTU are not available,
we reconstruct surfaces 
using PSR at octree depth 10 to generate training data.
We trim PSR surfaces using SurfaceTrimmer 
software provided in PSR with trimming value 8. 
We randomly extract batches of points from each training scene to
train our network. Even though we use only 
6 scenes in training set, we achieve a high accuracy and good 
generalization capability.
Table \ref{tab:dtu_accu} gives the quantitative 
results on testing scenes in DTU dataset \footnote{
	We use scenes \{1, 2, 3, 4, 5, 6\} as training set, scenes \{7, 8, 9\} 
	as validation set and scenes \{10, 11, 12, 13, 14, 15, 16, 18, 19, 
	21, 24, 29, 30, 34, 35, 38, 41, 42, 45, 46, 48, 51, 55, 
	59, 60, 61, 62, 63, 65, 69, 84, 93, 94, 95, 97, 
	106, 110, 114, 122, 126, 127, 128\} as testing set. 
}. 
Qualitative results are shown in 
Figure \ref{fig:dtu_quality} (a).

The point clouds in DTU are all open scenes. 
Although ONet can finish reconstruction tasks of watertight surfaces, 
it cannot reconstruct surface from point clouds of open scenes. 
Here we set results of PSR as an evaluation reference.
Surfaces reconstructed by PSR for evaluation
are also reconstructed at octree depth 9.	
The results of PSR are always closed surfaces even reconstructed from open 
scenes. Therefore, we trim them with trimming value 8 and 9.5.

As illustrated in Table \ref{tab:classification_accuracy_of_SSRNet_on_different_scale_datasets},
our network gets a 
high average classification accuracy 
of $ 95.7\% $.
As Table \ref{tab:dtu_accu} shows, compared with PSR with trimming value 8, our surfaces
perform better on DTU Accuracy, DTU Completeness and
CD. As for PSR with trimming value 9.5, we get similar 
performance on DTU Accuracy and perform better on
DTU Completeness. 
The PSR with trimming value 9.5 is more accurate on CD
while it is at cost of completeness. 
Our method differs from PSR that our method
directly produces open surfaces and we do not need to apply a
trimming step for our results, so we get more accurate and
complete borders than PSR (See Figure \ref{fig:dtu_quality} (a)).
In conclusion, 
the quality of
our method is comparable with PSR on DTU datasets. 
We perform 
better with respect to the completeness
on open scenes.
Detailed figures and videos are 
also provided in our
supplementary materials.

\subsection{Generalization Capability}\label{sec:generalizationability}		
In this section we evaluate the 
generalization capability of our method. It is worth noting that three datasets (ShapeNet, DTU, Stanford 3D) are quite different in 
terms of data scale and data category. 
Data in the ShapeNet are all synthetic.
The DTU only contains real-world scan data of open scenes
while the Stanford 3D only has closed scenes.
Therefore, tests among these three datasets 
can well evaluate the generalization performance of SSRNet. 

We test two trained models (SSRNet and ONet trained on ShapeNet) on Stanford 3D. 
Table \ref{tab:classification_accuracy_of_SSRNet_on_different_scale_datasets} and Table \ref{tab:generalization_on_stanford} 
show that 
we achieve high average 
classification accuracy of  $ 98.1\% $.
SSRNet generalizes better than ONet on Stanford 3D in 
metric of Chamfer Distance. 
Figure \ref{fig:dtu_quality} (b)
shows that 
our results achieve outstanding visual effects
and reconstruct geometry details of the shapes well. 

We also use SSRNet model trained on DTU
to test on Stanford 3D. 
 Table \ref{tab:generalization_on_stanford}
shows that our network trained on open scenes can also 
achieve higher classification accuracy 
on closed scenes. Besides, classification accuracy of 
our models trained on ShapeNet and DTU are almost exactly the same,
and the evaluation results on CD are also similar.
In general, the results prove good generalization
capability of our network. One strong benefit is 
that we do not need to retrain our network on 
different datasets to complete the 
reconstruction work.

\subsection{Efficiency}

We test our network on 4 GeForce GTX 1080 Ti GPUs in a  
Intel Xeon(R) CPU system with 32$ \times $2.10 GHz cores.
We implement data preprocessing method
with CUDA. 
In our experiments, we set the maximum number of point and 
vertex in one batch as 300,000.
As is illustrated in Table \ref{tab:total_time},
our method can tackle millions
of points and vertices in reasonable time \footnote{DTU \cite{jensen2014large} contains a series of point clouds, 
	here we just randomly list time consumption of stl\_030, stl\_034, stl\_062. We do not consider the time of reading and writing files.}.
As a learning-based method, 
the time consumption  
is generally acceptable.
Since our network is able to perform predictions parallelly using 
multi-GPUs, our prediction procedure can be accelerated
by more GPUs.
As shown in the table, 
the predictions with 4 GPUs are about 
2.6 times faster than 1 GPU.
This reveals the efficiency potential of
our method scaling to large datasets.
When conditions permitting (more GPUs), 
it can be accelerated a lot.


\section{Conclusion}

Our SSRNet has many advantages. Firstly, 
it has strong scalability
that allows dividing the input data
and processing different parts parallelly. 
Meanwhile, the network has also proved its efficiency by experienments in which it reconstructs quality surfaces from point clouds with millions of points in reasonable time.
Besides, SSRNet is 
good at reconstructing geometry details with the help of  
the geometry-aware features. 
These features are only related to 
local neighbor information
and do not rely on any global shape information.
Paying too much attention to global shape feature rather 
than just considering local geometry information would limit 
the generalization capability among different shape types.
%
%
Thus SSRNet also has good generalization capability
and it does not need too much training data, 
which avoids overfitting.

In conclusion, we have successfully 
designed a scalable network for quality surface 
reconstruction from point clouds and 
make a significant breakthrough 
compared with existing state-of-the-art learning-based methods.
We believe it would be an important step to make learning-based methods competitive with respect to geometry processing  algorithms on real-world and challenging data.

\section*{Acknowledgements}
 This work was supported by the National Natural Science Foundation of China under Grant 61772213, Grant 61991412 and Grant 91748204.

{\small
\bibliographystyle{ieee_fullname}
\bibliography{egbib}
}

\end{document}